\documentclass[10pt,twocolumn,letterpaper]{article}

\usepackage{cvpr}
\usepackage{times}
\usepackage{epsfig}
\usepackage{graphicx}
\usepackage{amsmath}
\usepackage{amssymb}
\usepackage{algorithm}
\usepackage{algpseudocode}
\usepackage{authblk}
\usepackage{url}
\usepackage[breaklinks=true,bookmarks=false]{hyperref}

\newcommand{\Eref}[1]{Equation~(\ref{#1})}
\newcommand{\Fref}[1]{Figure~\ref{#1}}
\newcommand{\Sref}[1]{Section~\ref{#1}}

\newcommand{\fref}[1]{Fig.~\ref{#1}}

\cvprfinalcopy % *** Uncomment this line for the final submission

\def\cvprPaperID{****} % *** Enter the CVPR Paper ID here

\graphicspath{ {images/} }
\begin{document}

\title{StyleBank: An Explicit Representation for Neural Image Style Transfer }

\author[1]{Dongdong Chen\thanks{This work was done when Dongdong Chen is an intern at MSR Asia.}}
\author[2]{Lu Yuan}
\author[2]{Jing Liao}
\author[1]{Nenghai Yu}
\author[2]{Gang Hua}
\affil[1]{University of Science and Technology of China, \authorcr \tt\small {cd722522@mail.ustc.edu.cn, ynh@ustc.edu.cn}}
\affil[2]{Microsoft Research Asia, \authorcr \tt\small{\{luyuan,jliao,ganghua\}@microsoft.com }}

\maketitle

%%%%%%%%% ABSTRACT
	\begin{abstract}

We propose \emph{StyleBank}, which is composed of multiple convolution filter banks and each filter bank explicitly represents one style, for neural image style transfer. To transfer an image to a specific style, the corresponding filter bank is operated on top of the intermediate feature embedding produced by a single auto-encoder. The StyleBank and the auto-encoder are jointly learnt, where the learning is conducted in such a way that the auto-encoder does not encode any style information thanks to the flexibility introduced by the explicit filter bank representation. It also enables us to conduct incremental learning to add a new image style by learning a new filter bank while holding the auto-encoder fixed. The explicit style representation along with the flexible network design enables us to fuse styles at not only the image level, but also the region level. Our method is the first style transfer network that links back to traditional texton mapping methods, and hence provides new understanding on neural style transfer. Our method is easy to train, runs in real-time, and produces results that qualitatively better or at least comparable to existing methods.

\end{abstract}

%%%%%%%%%%%%%%%%%%%%%%%%%%%%%%%%%%%%%%%%%%%%%%%%%%%%%%%%%%%%%%%%%%%%%%%%%%%%%%
%%%%%%%%% Introduction
%%%%%%%%%%%%%%%%%%%%%%%%%%%%%%%%%%%%%%%%%%%%%%%%%%%%%%%%%%%%%%%%%%%%%%%%%%%%%%

\section{Introduction}

Style transfer is to migrate a style from an image to another, and is closely related to texture synthesis. The core problem behind these two tasks is to model the statistics of a reference image (texture, or style image), which enables further sampling from it under certain constraints. For texture synthesis, the constraints are that the boundaries between two neighboring samples must have a smooth transition, while for style transfer, the constraints are that the samples should match the local structure of the content image. So in this sense, style transfer can be regarded as a generalization of texture synthesis.

Recent work on style transfer adopting Convolutional Neural Networks (CNN) ignited a renewed interest in this problem. On the machine learning side, it has been shown that a pre-trained image classifier can be used as a feature extractor to drive texture synthesis~\cite{gatys2015texture} and style transfer~\cite{gatys2015neural}. These CNN algorithms either apply an iterative optimization mechanism~\cite{gatys2015neural}, or directly learn a feed-forward generator network~\cite{johnson2016perceptual,ulyanov2016texture} to seek an image close to both the content image and the style image -- all measured in the CNN ({\em i.e.}, pre-trained VGG-16~\cite{simonyan2014very}) feature domain. These algorithms often produce more impressive results compared to the texture-synthesis ones, since the rich feature representation that a deep network can produce from an image would allow more flexible manipulation of an image.

Notwithstanding their demonstrated success, the principles of CNN style transfer are vaguely understood. After a careful examination of existing style transfer networks, we argue that the content and style are still coupled in their learnt network structures and hyper-parameters. To the best of our knowledge, an explicit representation for either style or content has not yet been proposed in these previous neural style transfer methods.

 As a result, the network is only able to capture a specific style one at a time. For a new style, the whole network has to be retrained end-to-end. In practice, this makes these methods unable to scale to large number of styles, especially when the style set needs to be incrementally augmented. In addition, how to further reduce run time, network model size and enable more flexibilities to control transfer (\eg., region-specific transfer), remain to be challenges yet to be addressed.

To explore an explicit representation for style, we reconsider neural style transfer by linking back to traditional texton (known as the basic element of texture) mapping methods, where mapping a texton to the target location is equivalent to a convolution between a texton and a Delta function (indicating sampling positions) in the image space.

Inspired by this, we propose \emph{StyleBank}, which is composed of multiple convolution filter banks and each filter bank represents one style. To transfer an image to a specific style, the corresponding filter bank is convolved with the intermediate feature embedding produced by a single auto-encoder, which decomposes the original image into multiple feature response maps. This way, for the first time, we provide a clear understanding of the mechanism underneath neural style transfer.

The \emph{StyleBank} and the auto-encoder are jointly learnt in our proposed feed-forward network. It not only allows us to simultaneously learn a bundle of various styles, but also enables a very efficient incremental learning for a new image style. This is achieved by learning a new filter bank while holding the auto-encoder fixed.

We believe this is a very useful functionality to recently emerged style transfer mobile applications (\eg, Prisma) since we do not need to train and prepare a complete network for every style. More importantly, it can even allow users to efficiently create their own style models and conveniently share to others. Since the part of our image encoding is shared for variant styles, it may provide a faster and more convenient switch for users between different style models.

Because of the explicit representation, we can more conveniently control style transfer and create new interesting style fusion effects. More specifically, we can either linearly fuse different styles altogether, or produce region-specific style fusion effects. In other words, we may produce an artistic work with hybrid elements from van Gogh's and Picaso's paintings.

Compared with existing neural style transfer networks ~\cite{johnson2016perceptual,ulyanov2016texture}, our proposed neural style transfer network is unique in the following aspects:
\begin{itemize}
\item{ In our method, we provide an explicit representation for styles. This enables our network to completely decouple styles from the content after learning.}

\item{Due to the explicit style representation, our method enables region-based style transfer. This is infeasible in existing neural style transfer networks, although classical texture transfer methods were able to achieve it. }

\item{Our method not only allows to simultaneously train multiple styles sharing a single auto-encoder, but also incrementally learn a new style without changing the auto-encoder. }

\end{itemize}

The remainder of the paper is organized as follows. We summarize related work in Section 2. We devote Section 3 to the main technical design of the proposed Style-Bank Network. Section 4 discusses about new characteristics of the proposed Style-Bank Network when compared with previous work. We present experimental results and comparisons in Section 5. And finally we conclude in Section 6.
	%%%%%%%%%%%%%%%%%%%%%%%%%%%%%%%%%%%%%%%%%%%%%%%%%%%%%%%%%%%%%%%%%%%%%%%%%%%%%%
%%%%%%%%% Related Work
%%%%%%%%%%%%%%%%%%%%%%%%%%%%%%%%%%%%%%%%%%%%%%%%%%%%%%%%%%%%%%%%%%%%%%%%%%%%%%

\section{Related Work}

Style transfer is very related to texture synthesis, which attempts to grow textures using non-parametric sampling of pixels~\cite{efros1999texture,wei2000fast} or patches~\cite{efros2001image,liang2001real} in a given source texture. The task of style transfer can be regarded as a problem of texture transfer~\cite{efros2001image,frigo2016split,elad2016style}, which synthesizes a texture from a source image constrained by the content of a target image. Hertzman et al.~\cite{hertzmann2001image} further introduce the concept of image analogies, which transfer the texture from an already stylised image onto a target image. However, these methods only use low-level image features of the target image to inform the texture transfer.

Ideally, a style transfer algorithms should be able to extract and represent the semantic image content from the target image and then render the content in the style of the source image. To generally separate content from style in natural images is still an extremely difficult problem before, but the problem is better mitigated by the recent development of Deep Convolutional Neural Networks (CNN)~\cite{krizhevsky2012imagenet}.

DeepDream~\cite{alexander2015deepdream} may be the first attempt to generate artistic work using CNN. Inspired by this work, Gatys et al.~\cite{gatys2015neural} successfully applies CNN (pre-trained VGG-16 networks) to neural style transfer and produces more impressive stylization results compared to classic texture transfer methods. This idea is further extended to portrait painting style transfer~\cite{selim2016painting} and patch-based style transfer by combining Markov Random Field (MRF) and CNN~\cite{chuanli2016}. Unfortunately, these methods based on an iterative optimization mechanism are computationally expensive in run-time, which imposes a big limitation in real applications.

To make the run-time more efficient, more and more works begin to directly learn a feed-forward generator network for a specific style. This way, stylized results can be obtained just with a forward pass, which is hundreds of times faster than iterative optimization~\cite{gatys2015neural}. For example, Ulyanov et al.~\cite{ulyanov2016texture} propose a texture network for both texture synthesis and style transfer. Johnson et al.~\cite{johnson2016perceptual} define a perceptual loss function to help learn a transfer network that aims to produce results approaching~\cite{gatys2015neural}. Chuan et al.~\cite{li2016precomputed} introduce a Markovian Generative Adversarial Networks, aiming to speed up their previous work~\cite{chuanli2016}.

However, in all of these methods, the learnt feed-forward networks can only represent one specific style. For a new style, the whole network has to be retrained, which may limit the scalability of adding more styles on demand. In contrast, our network allows a single network to simultaneously learn numerous styles. Moreover, our work enables incremental training for new styles.

At the core of our network, the proposed \emph{StyleBank} represents each style by a convolution filter bank. It is very analogous to the concept of "texton"~\cite{malik1999textons,zhu2005textons,li2002motion} and filter bank in \cite{zhu1998filters,lu2015learning}, but \emph{StyleBank} is defined in feature embedding space produced by auto-encoder~\cite{HinSal06} rather than image space. As we known, embedding space can provide compact and descriptive representation for original data~\cite{bengio2003neural,Scott2015analogy,xue2016visual}. Therefore, our \emph{StyleBank} would provide a better representation for style data compared to predefined dictionaries (such as wavelet~\cite{portilla2000parametric} or pyramid~\cite{heeger1995pyramid} ).

\begin{figure*}[ht]
	\centering \includegraphics[width=0.975\textwidth]{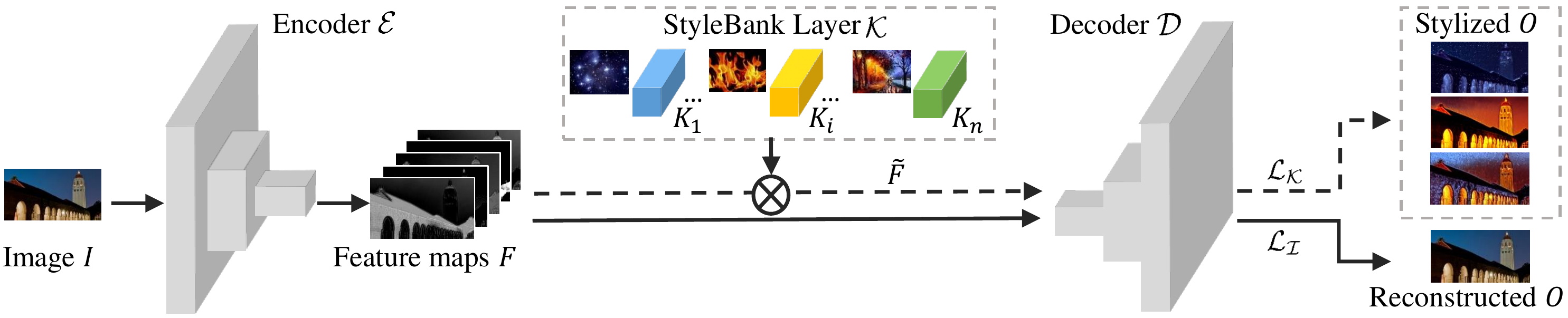}
	\caption{Our network architecture consists of three modules: image encoder $\mathcal{E}$, StyleBank layer $\mathcal{K}$ and image decoder $\mathcal{D}$}
	\label{fg:architecture}
\end{figure*}

\section{StyleBank Networks}

%%%%%%%%%%%%%%%%%%%%%%%%%%%%%%%%%%%%%%%%%%%%%%%%%%%%%%%%%%%%%%%%%%%%%%%%%%%%%%
%%%%%%%%% Method
%%%%%%%%%%%%%%%%%%%%%%%%%%%%%%%%%%%%%%%%%%%%%%%%%%%%%%%%%%%%%%%%%%%%%%%%%%%%%%

\subsection{StyleBank}

At its core, the task of neural style transfer requires a more explicit representation, like texton~\cite{malik1999textons,li2002motion} (known as the basic element of texture) used in classical texture synthesis. It may provide a new understanding for the style transfer task, and then help design a more elegant architecture to resolve the coupling issue in existing transfer networks~\cite{johnson2016perceptual,ulyanov2016texture}, which have to retrain hyper-parameters of the whole network for each newly added style end-to-end.

We build a feed-forward network based on a simple image auto-encoder (shown in~\Fref{fg:architecture}), which would first transform the input image (\ie, the \emph{content} image) into the feature space through the encoder subnetwork. Inspired by the texton concept, we introduce \emph{StyleBank} as style representation by analogy, which is learnt from input styles.

Indeed, our StyleBank contains multiple convolution filter banks. Every filter bank represents one kind of style, and all channels in a filter bank can be regarded as bases of style elements (\eg, texture pattern, coarsening or softening strokes). By convolving with the intermediate feature maps of content image, produced by auto-encoder, \emph{StyleBank} would be mapped to the content image to produce different stylization results. Actually, this manner is analogy to texton mapping in image space, which can also be interpreted as the convolution between texton and Delta function (indicating sampling positions).

\subsection{Network Architecture}

\Fref{fg:architecture} shows our network architecture, which consists of three modules: image encoder $\mathcal{E}$, \emph{StyleBank} layer $\mathcal{K}$ and image decoder $\mathcal{D}$, which constitute two learning branches: auto-encoder (\ie, $\mathcal{E}\rightarrow\mathcal{D}$) and stylizing (\ie, $\mathcal{E}\rightarrow\mathcal{K}\rightarrow\mathcal{D}$). Both branches share the same encoder $\mathcal{E}$ and decoder $\mathcal{D}$ modules.

Our network requires the \emph{content} image $\mathit{I}$ to be the input. Then the image is transformed into multi-layer feature maps $\mathit{F}$ through the encoder $\mathcal{E}$: $\mathit{F} = \mathcal{E}(\mathit{I})$. For the auto-encoder branch, we train the auto-encoder to produce an image that is as close as possible to the input image, \ie, $\mathit{O}=\mathcal{D}(\mathit{F}) \to \mathit{I}$. In parallel, for the stylizing branch, we add an intermediate \emph{StyleBank} layer $\mathcal{K}$ between $\mathcal{E}$ and $\mathcal{D}$. In this layer, \emph{StyleBank} $\{K_i\}, (i= 1, 2,...,n)$, for $n$ styles would be respectively convolved with features $\mathit{F}$ to obtain transferred features $\mathit{\widetilde{F}}_i$. Finally, the stylization result $\mathit{O}_i$ for style $i$ is achieved by the decoder $\mathcal{D}$: $\mathit{O}_i = \mathcal{D}(\mathit{\widetilde{F}}_i)$.

 In this manner, contents could be encoded to the auto-encoder $\mathcal{E}$ and $\mathcal{D}$ as much as possible, while styles would be encoded into \emph{StyleBank}. As a result, content and style are decoupled from our network as much as possible.

\paragraph{Encoder and Decoder.} Following the architecture used in~\cite{johnson2016perceptual}, the image encoder $\mathcal{E}$ consists of one stride-1 convolution layer and two stride-2 convolution layers, symmetrically, the image decoder $\mathcal{D}$ consists of two stride-$\frac{1}{2}$ fractionally strided convolution layers and one stride-1 convolution layer. All convolutional layers are followed by instance normalization~\cite{ulyanov2016instance} and a ReLU nolinearity except the last output layer. Instance normalization has been demonstrated to perform better than spatial batch normalization~\cite{ioffe2015batch} in handling boundary artifacts brought by padding. Other than the first and last layers which use $9 \times 9$ kernels, all convolutional layers use $3 \times 3$ kernels. Benefited from the explicit representation, our network can remove all the residual blocks~\cite{he2015deep} used in the network presented in Johnson et al.~\cite{johnson2016perceptual} to further reduce the model size and computation cost without performance degradation.

\paragraph{StyleBank Layer.} Our architecture allows multiple styles (by default, 50 styles, but there is really no limit on it) to be simultaneously trained in the single network at the beginning. In the \emph{StyleBank} layer $\mathcal{K}$, we learn $n$ convolution filter banks $\{K_i\}, (i = 1, 2, ...n)$ (referred as \emph{StyleBank}). During training, we need to specify the $i$-th style, and use the corresponding filter bank $K_i$ for forward and backward propagation of gradients. At this time, transferred features $\mathit{\widetilde{F}}_i$ is achieved by
\begin{equation}\label{eq:style_embed}
\mathit{\widetilde{F}}_i = \mathit{K}_{i} \otimes \mathit{F},
\end{equation}
where $\mathit{F} \in \mathcal{R}^{c_{in} \times h \times w}$, $\mathit{K}_{i} \in \mathcal{R}^{c_{out} \times c_{in} \times k_h \times k_w}$, $\mathit{\widetilde{F}} \in \mathcal{R}^{c_{out} \times h \times w}$, $c_{in}$ and $c_{out}$ are numbers of feature channels for $\mathit{F}$ and $\mathit{\widetilde{F}}$ respectively, $(h, w)$ is the feature map size, and $(k_w, k_h)$ is the kernel size. To allow efficient training of new styles in our network, we may reuse the encoder $\mathcal{E}$ and the decoder $\mathcal{D}$ in our new training. We fix the trained $\mathcal{E}$ and $\mathcal{D}$, and only retrain the layer $\mathcal{K}$ with new filter banks starting from random initialization.
	
\paragraph{Loss Functions.} Our network consists of two branches: auto-encoder (\ie, $\mathcal{E}\rightarrow\mathcal{D}$) and stylizing (\ie, $\mathcal{E}\rightarrow\mathcal{K}\rightarrow\mathcal{D}$), which are alternatively trained. Thus, we need to define two loss functions respectively for the two branches.

 In the auto-encoder branch, we use MSE (Mean Square Error) between input image $I$ and output image $O$ to measure an \emph{identity loss} $\mathcal{L}_{\mathcal{I}}$:
\begin{equation}\label{eq:loss_identity}
\mathcal{L}_{\mathcal{I}}(\mathit{I},\mathit{O}) = \lVert\mathit{O}-\mathit{I}\rVert^{2}.
\end{equation}

At the stylizing branch, we use \emph{perceptual loss} $\mathcal{L}_{\mathcal{K}}$ proposed in~\cite{johnson2016perceptual}, which consists of a content loss $\mathcal{L}_{c}$, a style loss $\mathcal{L}_{s}$ and a variation regularization loss $\mathcal{L}_{tv}(O_i)$:
\begin{equation} \label{eq:loss_perceptual} \mathcal{L}_{\mathcal{K}}(\mathit{I},\mathit{S_i},\mathit{O_i})=\alpha\mathcal{L}_{c}(\mathit{O_i},\mathit{I})+\beta\mathcal{L}_{s}(\mathit{O_i},\mathit{S_i})+\gamma\mathcal{L}_{tv}(O_i)
\end{equation}
where $\mathit{I}$, $\mathit{S_i}$, $\mathit{O_i}$ are the input content image, style image and stylization result (for the $i$-th style) respectively. $\mathcal{L}_{tv}(O_i)$ is a variation regularizer used in~\cite{aly2005image,johnson2016perceptual}. $\mathcal{L}_{c}$ and $\mathcal{L}_{s}$ use the same definition in ~\cite{gatys2015neural}:
\begin{equation}\label{eq:loss_def}
\begin{aligned}
\mathcal{L}_{c}(\mathit{O_i}, \mathit{I}) &= \sum\mathop{}_{l \in \{l_c\}}\lVert F^l(O_i) - F^l(I)\rVert^2
\\
\mathcal{L}_{s}(\mathit{O_i}, \mathit{S}) &= \sum\mathop{}_{l \in \{l_s\}}\lVert G(F^l(O_i)) - G(F^l(S_i))\rVert^2
\end{aligned}
\end{equation}
where $\mathit{F}^l$ and $\mathit{G}$ are respectively feature map and Gram matrix computed from layer $l$ of VGG-16 network~\cite{simonyan2014very}(pre-trained on the ImageNet dataset ~\cite{russakovsky2015imagenet}). $\{l_c\}, \{l_s\}$ are VGG-16 layers used to respectively compute the content loss and the style loss.

\paragraph{Training Strategy.} \label{train_strategy}
We employ a $(T+1)$-step alternative training strategy motivated by~\cite{goodfellow2014generative} in order to balance the two branches (auto-encoder and stylizing). During training, for every $T+1$ iterations, we first train $T$ iterations on the branch with $\mathcal{K}$, then train one iteration for auto-encoder branch. We show the training process in Algorithm~\ref{ag:train_strategy}.
\begin{algorithm}
\caption{Two branches training strategy. Here $\lambda$ is the tradeoff between two branches. $\Delta_{\theta_{\mathcal{K}}}$ denote gradients of filter banks in $\mathcal{K}$. $\Delta_{\theta_{\mathcal{E,D}}}^\mathcal{K},\Delta_{\theta_{\mathcal{E,D}}}^\mathcal{I}$ denote gradients of $\mathcal{E,D}$ in stylizing and auto-encoder branches respectively.}\label{ag:train_strategy}
\begin{algorithmic}
\For {\;every $T+1$ iterations}
   \State {//\;Training at branch\; $\mathcal{E}\rightarrow\mathcal{K}\rightarrow\mathcal{D}$:}
    \For{$t=1$ to $T$}
        \State{$\mathord{\bullet}$ Sample $m$ images $\mathit{X}=\{x_i\}$ and style indices \\ \quad\quad\quad\;\;\;$\mathit{Y}=\{y_i\}$, $i \in \{1, ..., m\}$ as one mini-batch.}
        \State{$\mathord{\bullet}$ Update $\mathcal{E,D}$ and $\{K_j\}, j \in \mathit{Y}$:}
        \State {\qquad$\Delta_{\theta_{\mathcal{E,D}}}^{\mathcal{K}} \gets \bigtriangledown_{\theta_{\mathcal{E,D}}}\mathcal{L}_{\mathcal{K}}$}
        \State {\qquad$\Delta_{\theta_{\mathcal{K}}} \gets \bigtriangledown_{\theta_{\mathcal{K}}}\mathcal{L}_{\mathcal{K}}$}
    \EndFor{}
    \State{//\;Training at branch\; $\mathcal{E}\rightarrow\mathcal{D}$:}
    \State{$\mathord{\bullet}$ Update $\mathcal{E,D}$ only:}
    \State{$\qquad\Delta_{\theta_{\mathcal{E,D}}}^{\mathcal{I}} \gets \bigtriangledown_{\theta_{\mathcal{E,D}}}\mathcal{L}_{\mathcal{I}}$}
    \State{$\qquad\Delta_{\theta_{\mathcal{E,D}}}^{\mathcal{I}}\gets \lambda \frac{\lVert\Delta_{\theta_{\mathcal{E,D}}}^{\mathcal{K}}\rVert}{\lVert\Delta_{\theta_{\mathcal{E,D}}}^{\mathcal{I}}\rVert}\Delta_{\theta_{\mathcal{E,D}}}^{\mathcal{I}}$}
\EndFor{}
\end{algorithmic}
\end{algorithm}

\subsection{Understanding StyleBank and Auto-encoder} \label{sc:working_principle}

For our new representation of styles, there are several questions one might ask:

\noindent \textbf{1) \emph{How does StyleBank represent styles? } }\label{sc:understanding_stylebank}

\begin{figure*}[ht]	
    \centering	
	\includegraphics[width=0.95\textwidth]{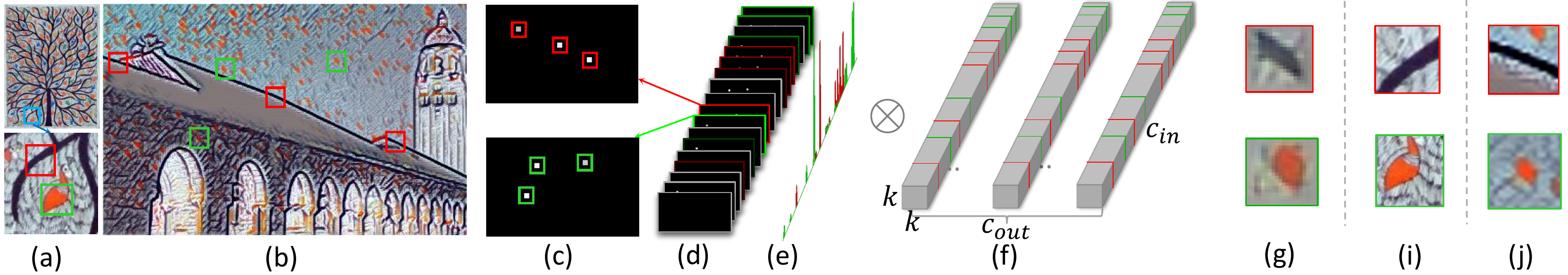}
	\caption{Reconstruction of the style elements learnt from two kinds of representative patches in an exemplar stylization image.}
    \label{fg:visualization}
\end{figure*}

After training the network, each styles is encoded in one convolution filter bank. Each channel of filter bank can be considered as dictionaries or bases in the literature of representation learning method~\cite{bengio2013representation}. Different weighted combinations of these filter channels can constitute various style elements, which would be the basic elements extracted from the style image for style synthesis. We may link them to ``textons" in texture synthesis by analogy.

For better understanding, we try to reconstruct style elements from a learnt filter bank in an exemplar stylization image shown in \Fref{fg:visualization}. We extract two kinds of representative patches from the stylization result (in \Fref{fg:visualization}(b))-- stroke patch (indicated by red box) and texture patch (indicated by green box) as an object to study. Then we apply two operations below to visualize what style elements are learnt in these two kinds of patches.

First, we mask out other regions but only remain these corresponding positions of the two patches in feature maps (as shown in \Fref{fg:visualization}(c)(d)), that would be convolved with the filter bank (corresponding to a specific style). We further plot feature responses in \Fref{fg:visualization}(e) for the two patches along the dimension of feature channels. As we can observe, their responses are actually sparsely distributed and some peak responses occur at individual channels. Then, we only consider non-zero feature channels for convolution and their convolved channels of filter bank (marked by green and red colors in \Fref{fg:visualization}(f)) indeed contribute to a certain style element. Transferred features are then passed to the decoder. Recovery style elements are shown in \Fref{fg:visualization}(g), which are very close in appearance to the original style patches (\Fref{fg:visualization}(i)) and stylization patches (\Fref{fg:visualization}(j)).

\begin{figure}
  \centering
  \includegraphics[width=0.48\textwidth]{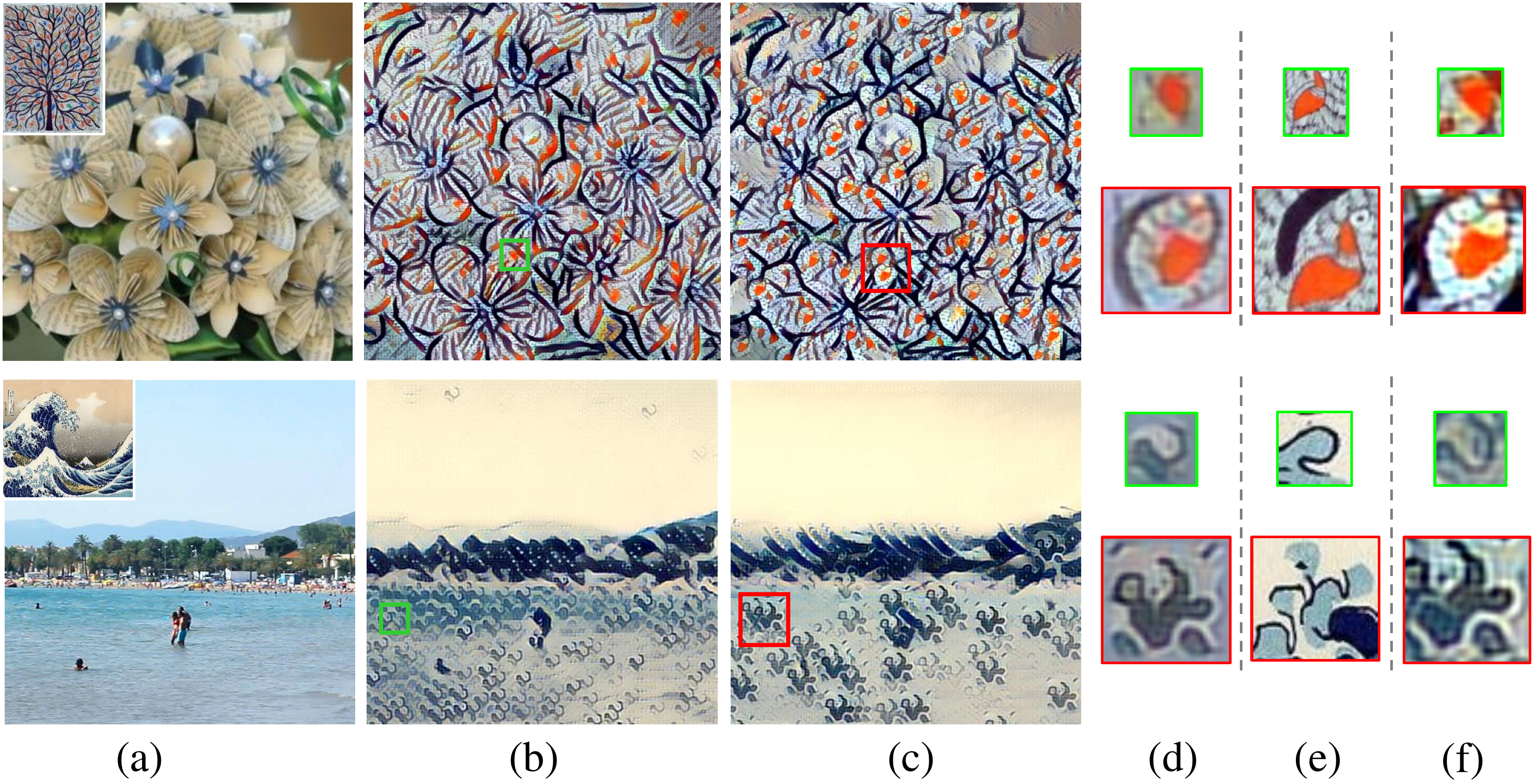}
  \caption{Learnt style elements of different StyleBank kernel sizes. (b) and (c) are stylization results of $(3,3)$ and $(7,7)$ kernels respectively. (d), (e) and (f) respectively show learnt style elements, original style patches and stylization patches.  }
  %It is easy to observe that bigger style elements could be learned if bigger $k$ is used
  \label{fg:ablation_kernelsize}
\end{figure}

To further explore the effect of kernel size $(k_w,k_h)$ in the StyleBank, we set a comparison experiment to train our network with two different kernel size of $(3,3)$ and $(7,7)$. Then we use similar method to visualize the learnt filter banks, as shown in \fref{fg:ablation_kernelsize}. Here the green and red box indicate representative patches from (3,3) and (7,7) kernels respectively. After comparison, it is easy to observe that bigger style elements can be learnt with larger kernel size. For example, in the bottom row , bigger sea spray appears in the stylization result with (7,7) kernels. That suggests our network supports the control on the style element size by tuning parameters to better characterize the example style.

\noindent \textbf{2) \emph{What is the content image encoded in?} }\label{sc:layer_decomposition}

\begin{figure}[t]
\centering
\includegraphics[width=0.48\textwidth]{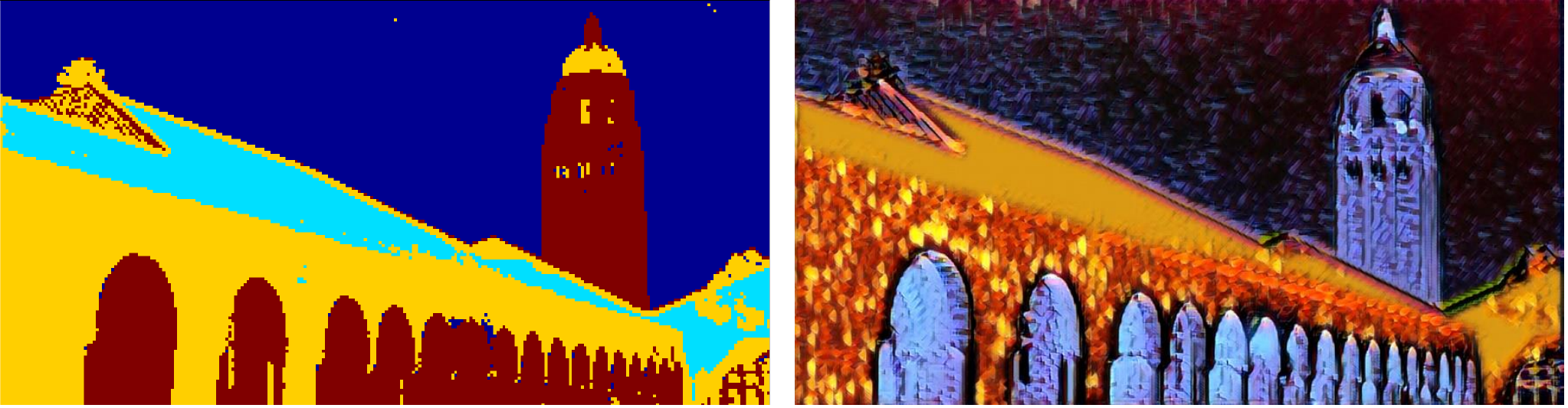}
\caption{k-means clustering result of feature maps(left) and corresponding stylization result(right).}
\label{fg:vis_layerwise}
\end{figure}

\begin{figure}[t]
\centering
\includegraphics[width=0.5\textwidth]{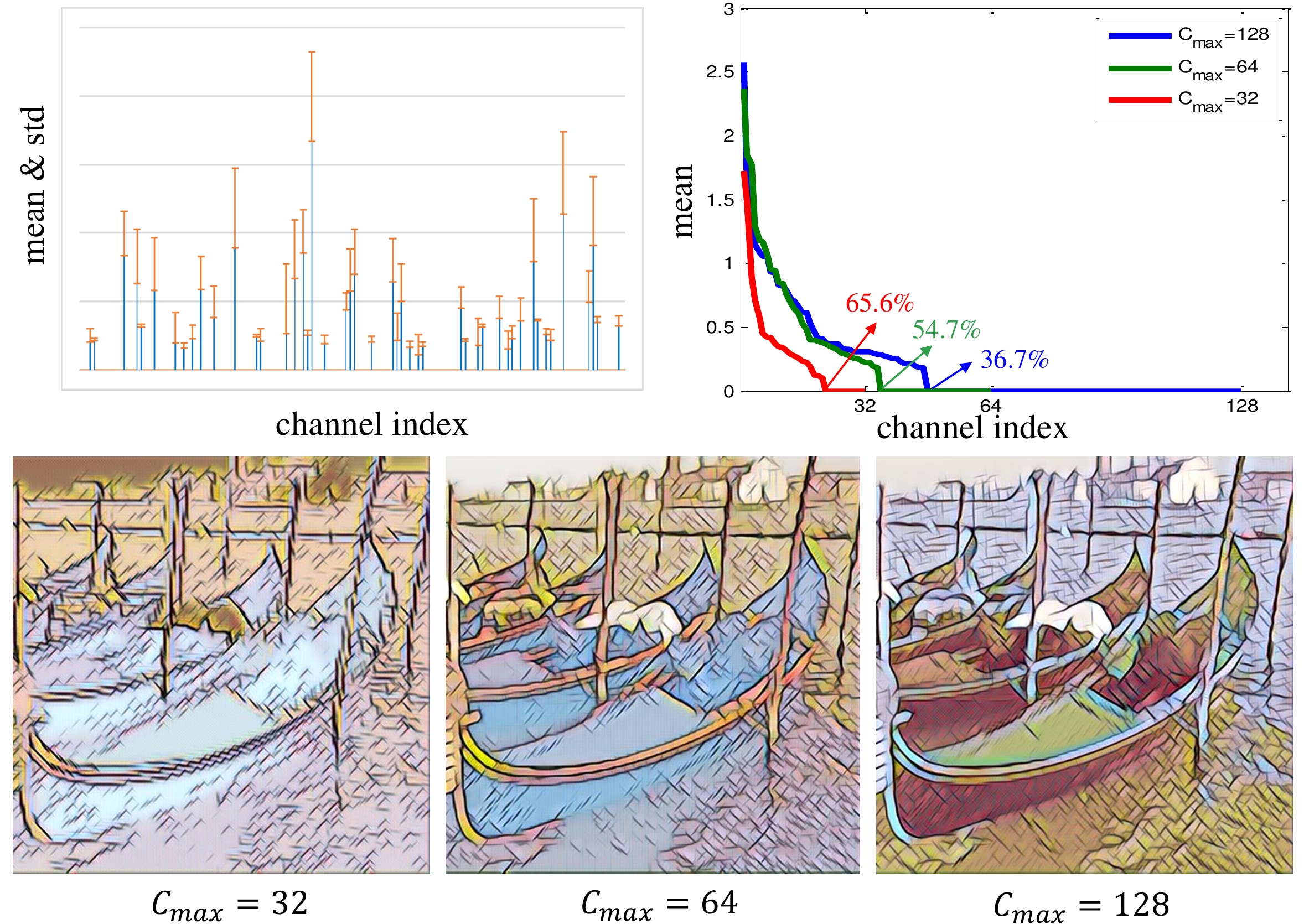}
\caption{Sparsity analysis. Top-left: means and standard deviations of per-channel average response; top-right: distributions of sorted means of per-channel average response for different model sizes ($C_{max} = 32,64,128$); bottom: corresponding stylization results.}
\label{fg:sparse_analysis}
\end{figure}

In our method, the auto-encoder is learnt to decompose the content image into multi-layer feature maps, which are independent of any styles. When further analyzing these feature maps, we have two observations.

First, these features can be spatially grouped into meaningful clusters in some sense (\eg, colors, edges, textures). To verify this point, we extract each feature vector at every position of feature maps. Then, an unsupervised clustering (\eg, K-means algorithms) is applied to all feature vectors (based on L2 normalized distance). Finally, we can obtain the clustering results shown in left of \Fref{fg:vis_layerwise}, which suggests a certain segmentation to the content image.

Comparing the right stylization result with left clustering results, we can easily find that different segmented regions are indeed rendered with different kinds of colors or textures. For regions with the same cluster label, the filled color or textures are almost the same. As a result, our auto-encoder may enable region-specific style transfer.
%and the stylization may be determined by the content of image regions.

Second, these features would distribute sparsely in channels. To exploit this point, we randomly sample $200$ content images, and for each image, we compute the average of all non-zero responses at every of $128$ feature channels (in the final layer of encoder). And then we plot the means and standard deviations of those per-channel averages among $200$ images in the top-left of~\Fref{fg:sparse_analysis}. As we can see, valuable responses consistently exist at certain channels. One possible reason is that these channels correspond to specific style elements for region-specific transfer, which is in consistency with our observation in~\Fref{fg:visualization}(e).

The above sparsity property will drive us to consider smaller model size of the network. We attempt to reduce all channel numbers in our auto-encoder and StyleBank layer by a factor of $2$ or $4$. Then the maximum channel number $C_{max}$ become 64, 32 respectively from the original 128. We also compute and sort the means of per-channel averages, as plotted in the top-right of \fref{fg:sparse_analysis}. We can observe that the final layer of our encoder still maintains the sparsity even for smaller models although sparsity is decreased in smaller models ($C_{max}=32$). On the bottom of~\Fref{fg:sparse_analysis}, we show corresponding stylization results of $C_{max}=32,64,128$ respectively. By comparison, we can notice that $C_{max}=32$ obviously produces worse results than $C_{max}=128$ since the latter may encourage better region decomposition for transfer. Nevertheless, there may still be a potential to design a more compact model for content and style representation. We leave that to our future exploration.

\noindent \textbf{3) \emph{How are content and style decoupled from each other?}}

\begin{figure}
	\centering
	\includegraphics[width=1.0\columnwidth]{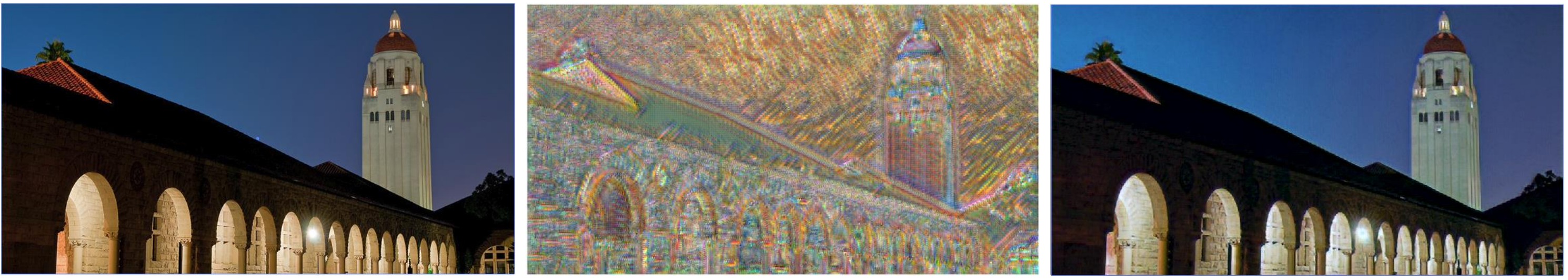}
	\caption{Illustration of the effects of two branches. The middle and right ones are reconstructed input image (left) with and without auto-encoder branch during training.}
    \label{fg:shortcut}
\end{figure}

To further know how well content is decoupled from style, we need to examine if the image is completely encoded in the auto-encoder. We compare two experiments with and without the auto-encoder branch in our training. When we only consider the stylizing branch, the decoded image (shown in the middle of ~\Fref{fg:shortcut}) produced by solely auto-encoder without $\mathcal{K}$ fails to reconstruct the original input image (shown in the left of ~\Fref{fg:shortcut}), and instead seems to carry some style information. When we enable the auto-encoder branch in training,  we obtain the final image (shown in the right of \Fref{fg:shortcut}) reconstructed from the auto-encoder, which has very close appearance to the input image. Consequently, the content is explicitly encoded into the auto-encoder, and independent of any styles. This is very convenient to carry multiple styles learning in a single network and reduce the interferences among different styles.

\noindent \textbf{4) \emph{How does the content image control style transfer?}}

\begin{figure}[t]
	\centering
	\includegraphics[width=0.48\textwidth]{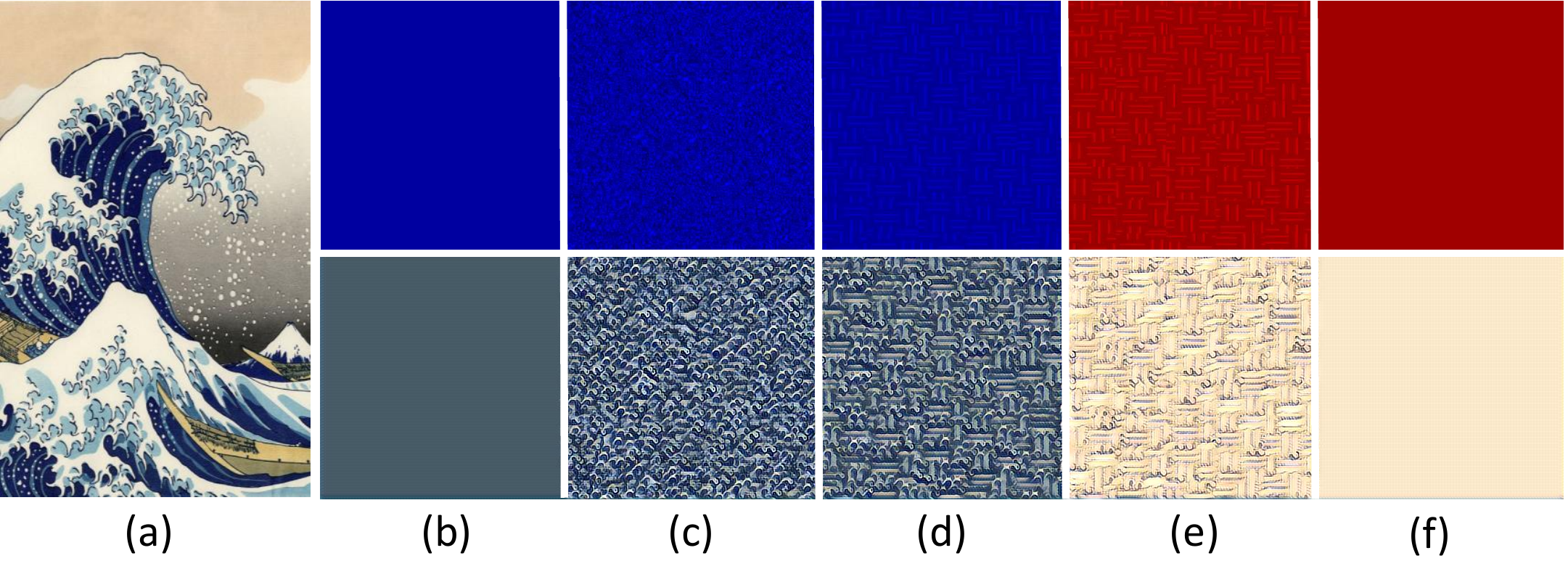}
	\caption{Stylization result of a toy image, which consists of four parts of different color or different texture. }
    \label{fg:vis_toy}
\end{figure}

To know how the content controls style transfer, we consider a toy case shown in \Fref{fg:vis_toy}. On the top, we show the input toy image consisting of five regions with variant colors or textures. On the bottom, we show the output stylization result. Below are some interesting observations:
\begin{itemize}
	\item For input regions with different colors but without textures, only a purely color transfer is applied (see \Fref{fg:vis_toy} (b)(f)).
	\item For input regions with the same color but different textures, the transfer consists of two parts: the same color transfer and different texture transfer influenced by appearance of input textures. (see \Fref{fg:vis_toy} (c)(d)).
	\item For input regions with different colors but the same textures, the results have the same transferred textures but different target colors (see \Fref{fg:vis_toy} (d)(e)).
\end{itemize}

%%%%%%%%%%%%%%%%%%%%%%%%%%%%%%%%%%%%%%%%%%%%%%%%%%%%%%%%%%%%%%%%%%%%%%%%%%%%%%
%%%%%%%%% Advantages of Our Method
%%%%%%%%%%%%%%%%%%%%%%%%%%%%%%%%%%%%%%%%%%%%%%%%%%%%%%%%%%%%%%%%%%%%%%%%%%%%%%
	
\section{Capabilities of Our Network}

Because of an explicit representation, our proposed feed-forward network provides additional capabilities, when compared with previous feedforward networks for style transfer. They may bring new user experiences or generate new stylization effects compared to existing methods.

\subsection{Incremental Training}

Previous style transfer networks (\eg,~\cite{johnson2016perceptual,ulyanov2016texture,chuanli2016}) have to be retrained for a new style, which is very inconvenient. In contrast, an iterative optimization mechanism~\cite{gatys2015neural} provides an online-learning for any new style, which would take several minutes for one style on GPU (\eg, Titan X). Our method has virtues of both feed-forward networks~\cite{johnson2016perceptual,ulyanov2016texture,chuanli2016} and iterative optimization method~\cite{gatys2015neural}. We enable an incremental training for new styles, which has comparable learning time to the online-learning method ~\cite{gatys2015neural}, while preserving efficiency of feed-forward networks~\cite{johnson2016perceptual,ulyanov2016texture,chuanli2016}.

In our configuration, we first jointly train the auto-encoder and multiple filter banks (50 styles used at the beginning) with the strategy described in Algorithm~\ref{ag:train_strategy}. After that, it allows to incrementally augment and train the \emph{StyleBank} layer for new styles by fixing the auto-encoder. The process converges very fast since only the augmented part of the \emph{StyleBank} would be updated in iterations instead of the whole network. In our experiments, when training with Titan X and given training image size of 512, it only takes around 8 minutes with about $1,000$ iterations to train a new style, which can speed up the training time by $20\sim40$ times compared with previous feed-forward methods.

\begin{figure}
	\centering
	\includegraphics[width=0.48\textwidth]{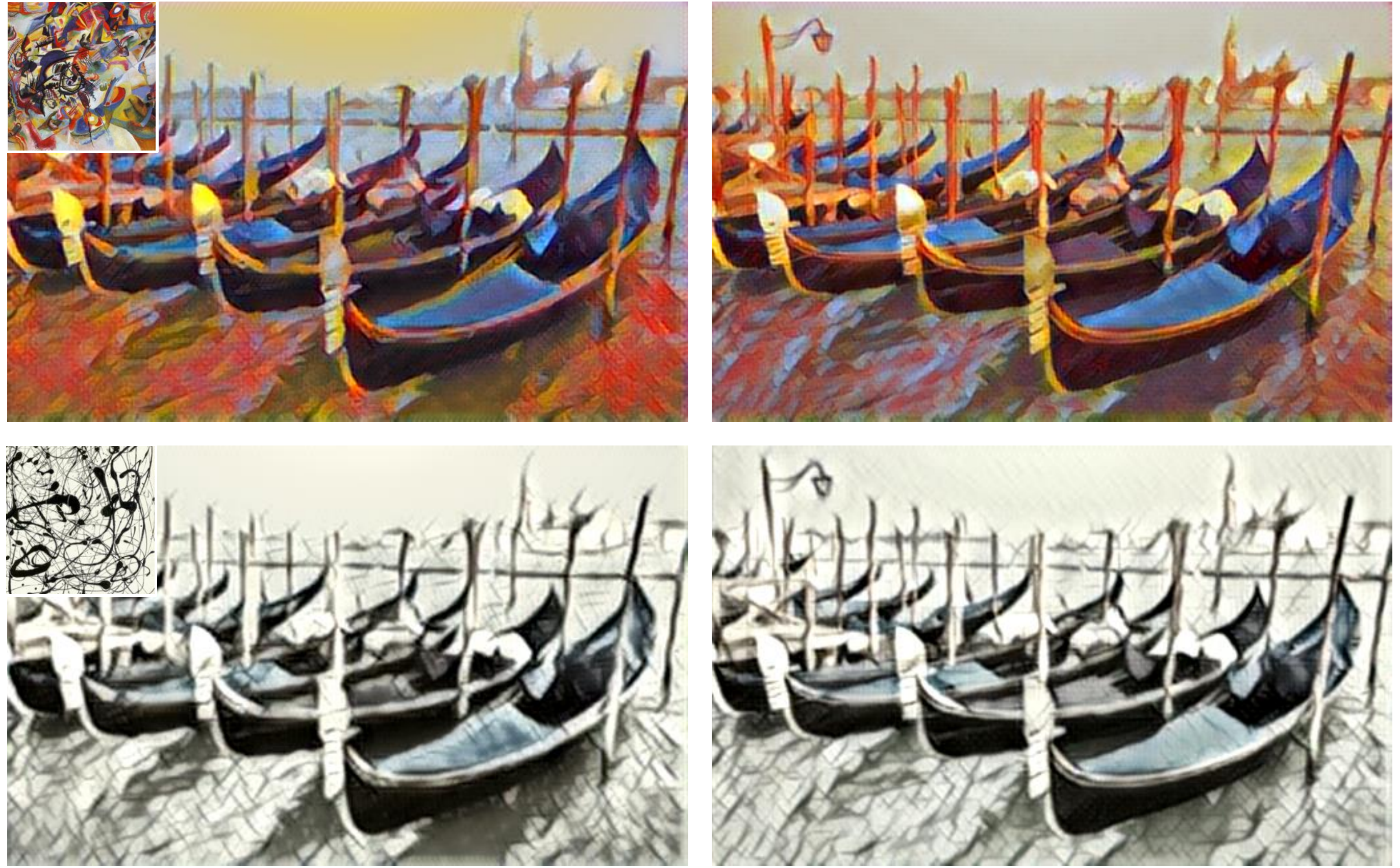}
	\caption{Comparison between incremental training (\emph{Left}) and fresh training (\emph{Right}). The target styles are shown on the top-left.}
	\label{fg:incremental_style_result}
\end{figure}

\Fref{fg:incremental_style_result} shows several stylization results of new styles by incremental training. It obtains very comparable stylization results to those from fresh training, which retrains the whole network with the new styles.

\subsection{Style Fusion}
We provide two different types of style fusion: linear fusion of multiple styles, and region-specific style fusion.

\begin{figure}
	\centering
	\includegraphics[width=0.48\textwidth]{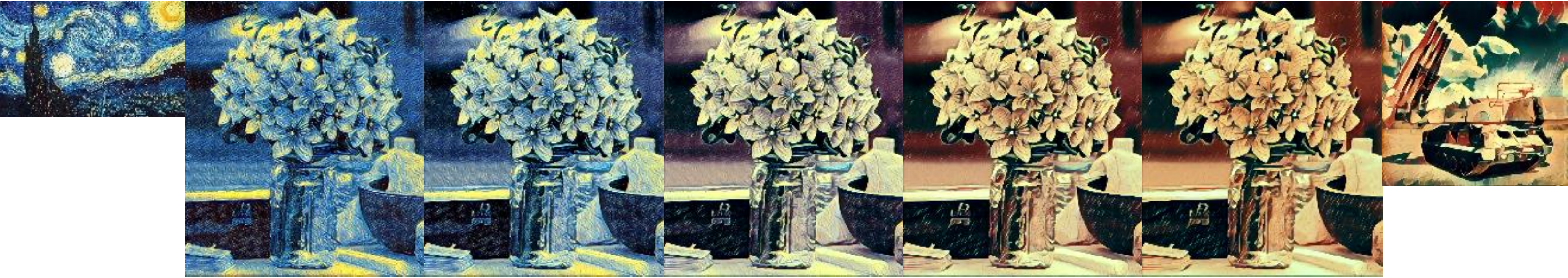}
	\caption{Results by linear combination of two style filter banks.}
    \label{fg:lc_stylefuse}
\end{figure}

\paragraph{Linear Fusion of Styles.} Since different styles are encoded into different filter banks $\{K_i\}$, we can linearly fuse multiple styles by simply linearly fusing filter banks in the \emph{StyleBank} layer. Next, the fused filter bank is used to convolve with content features $\mathit{F}$:
\begin{equation}\label{eq:lc_sf}
\begin{aligned}
\mathit{\widetilde{F}} &= (\sum\nolimits_{i=1}^{m}w_{i}*\mathit{K}_{i}) \otimes \mathit{F}\qquad \sum\nolimits_{i=1}^{m}w_{i} = 1,
\end{aligned}
\end{equation}
where $m$ is the number of styles, $\mathit{K}_{i}$ is the filter bank of style $i$. $\mathit{\widetilde{F}}$ is then fed to the decoder. \Fref{fg:lc_stylefuse} shows such linear fusion results of two styles with variant fusion weight $w_{i}$.
	
\paragraph{Region-specific Style Fusion.} Our method naturally allows a region-specific style transfer, in which different image regions can be rendered by various styles. Suppose that the image is decomposed into $n$ disjoint regions by automatic clustering (\eg, K-means mentioned in~\Sref{sc:working_principle} or advanced segmentation algorithms~\cite{boykov2001interactive,rother2004grabcut}) in our feature space, and $\mathit{M}_{i}$ denotes every region mask. The feature maps can be described as $\mathit{F} = \sum_{i=1}^{m}(\mathit{M}_{i}\times F)$. Then region-specific style fusion can be formulated as~\Eref{eq:rs_sf}:
\begin{equation}\label{eq:rs_sf}
\begin{aligned}
\mathit{\widetilde{F}} &= \sum\nolimits_{i=1}^{m}\mathit{K}_{i} \otimes (\mathit{M}_{i}\times F),
\end{aligned}
\end{equation}
where $\mathit{K}_{i}$ is the $i$-th filter bank.

\begin{figure}
	\centering
	\includegraphics[width=0.48\textwidth]{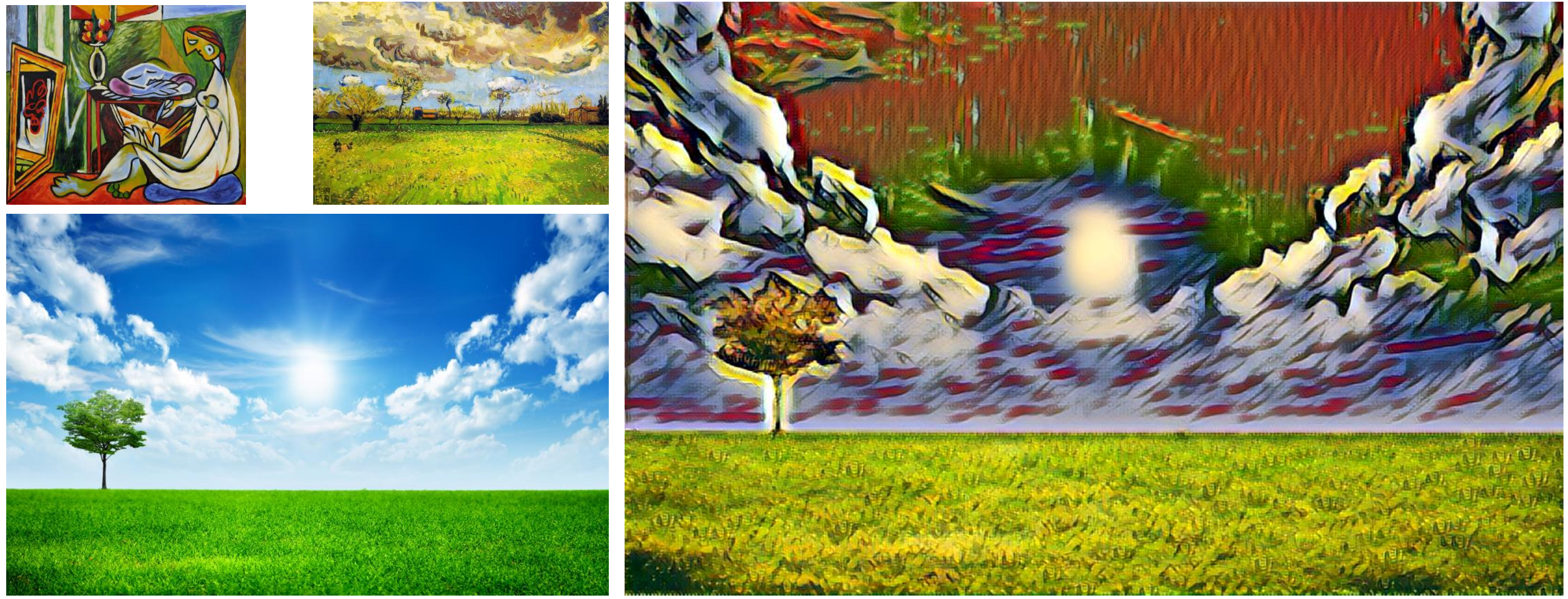}
	\caption{Region-specific style fusion with two paintings of Picasso and Van Gophm, where the regions are automatically segmented with K-means method.}
    \label{fg:rs_stylefuse}
\end{figure}

\Fref{fg:rs_stylefuse} shows such a region-specific style fusion result which exactly borrows styles from two famous paintings of Picasso and Van Goph. Superior to existing feed-forward networks, our method naturally obtains image decomposition for transferring specific styles, and passes the network only once. On the contrary, previous approaches have to pass the network several times and finally montage different styles via additional segmentation masks. %That is inconvenient and several times slower.

	%%%%%%%%%%%%%%%%%%%%%%%%%%%%%%%%%%%%%%%%%%%%%%%%%%%%%%%%%%%%%%%%%%%%%%%%%%%%%%
	%%%%%%%%% Experiments
	%%%%%%%%%%%%%%%%%%%%%%%%%%%%%%%%%%%%%%%%%%%%%%%%%%%%%%%%%%%%%%%%%%%%%%%%%%%%%%
	%------------------------------------------------------------------------
	\section{Experiments}
	\paragraph{Training Details}
	Our network is trained on 1000 content images randomly sampled from Microsoft COCO dataset~\cite{lin2014microsoft} and $50$ style images (from existing papers and the Internet). Each content image is randomly cropped to $512 \times 512$, and each style image is scaled to $600$ on the long side. We train the network with a batch size of 4 ($m=4$ in Algorithm~\ref{ag:train_strategy}) for $300k$ iterations. And the Adam optimization method~\cite{kingma2014adam} is adopted with the initial learning rate of $0.01$ and decayed by $0.8$ at every $30k$ iterations. In all of our experiments, we compute content loss at layer $relu4\_2$ and style loss at layer $relu1\_2$, $relu2\_2$, $relu3\_2$, and $relu4\_2$ of the pre-trained VGG-16 network. We use $T=2, \lambda = 1$ (in Algorithm~\ref{ag:train_strategy}) in our two branches training.

\subsection{Comparisons}

In this section, we compare our method with other CNN-based style transfer approaches~\cite{gatys2015neural,johnson2016perceptual,ulyanov2016texture,2016arXiv161007629D}. For fair comparison, we directly borrow results from their papers. It is difficult to compare results with different abstract stylization, which is indeed controlled by the ratio $\alpha/\beta$ in~\Eref{eq:loss_perceptual} and different work may use their own ratios to present results. For comparable perception quality, we choose different $\alpha,\beta$ in each comparison. More results are available in our supplementary material\footnote{\url{http://home.ustc.edu.cn/~cd722522/}}.

\paragraph{Compared with the Iterative Optimization Method.}
\begin{figure}
		\centering
		\includegraphics[width=\columnwidth]{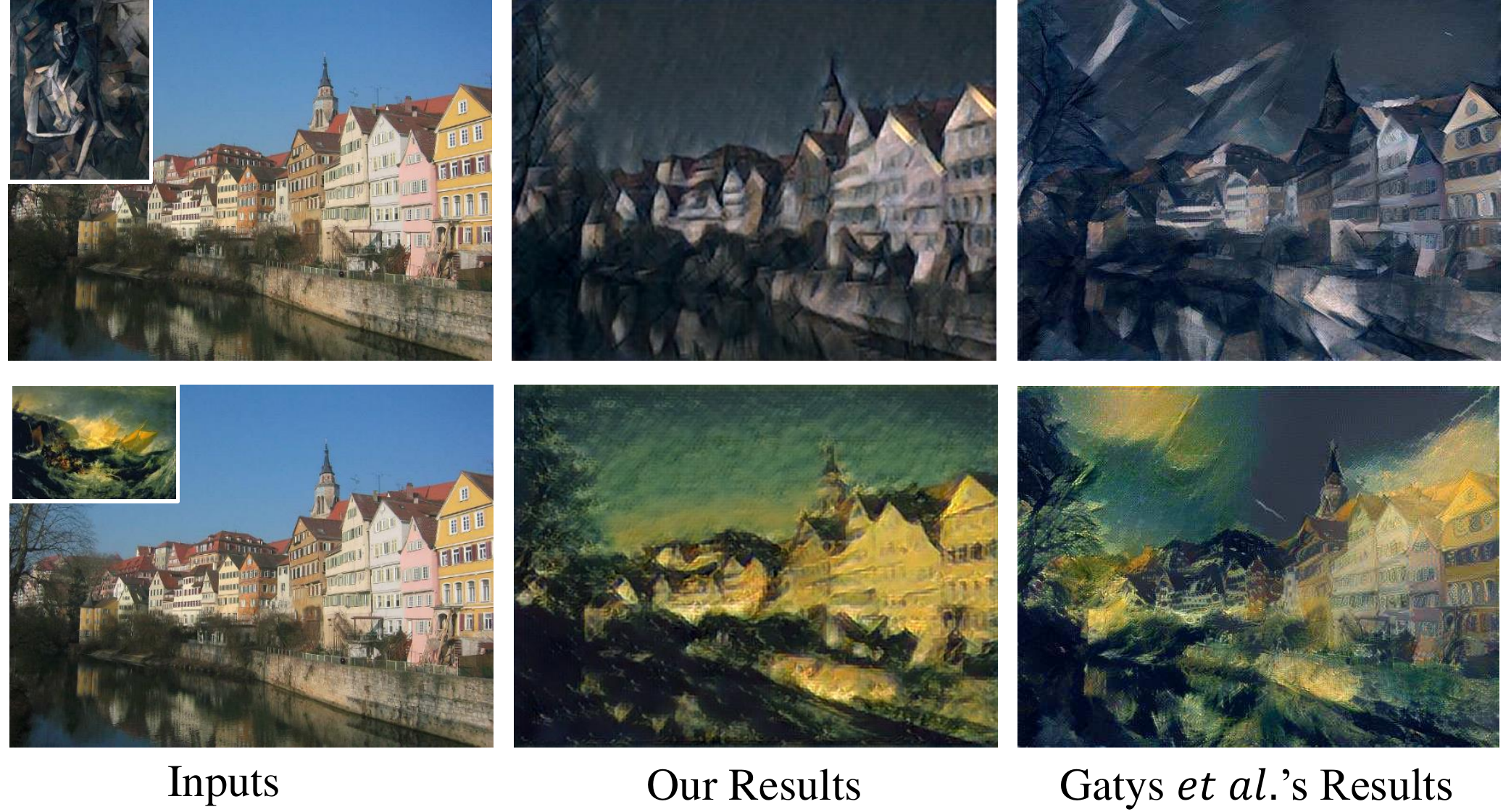}
		\caption{Comparison with optimization-based method \cite{gatys2015neural}. }
        \label{fg:comparison_neuralstyle}
\end{figure}

We use $\alpha/\beta = 1/100$ (in~\Eref{eq:loss_perceptual}) to produce comparable perceptual stylization in \fref{fg:comparison_neuralstyle}. Our method, like all other feed-forward methods, creates less abstract stylization results than optimization method \cite{gatys2015neural}. It is still difficult to judge which one is more appealing in practices. However, our method, like other feed-forward methods, could be hundreds of times faster than optimization-based methods.

\paragraph{Compared with Feed-forward Networks.}
\begin{figure}
		\centering
		\includegraphics[width=\columnwidth]{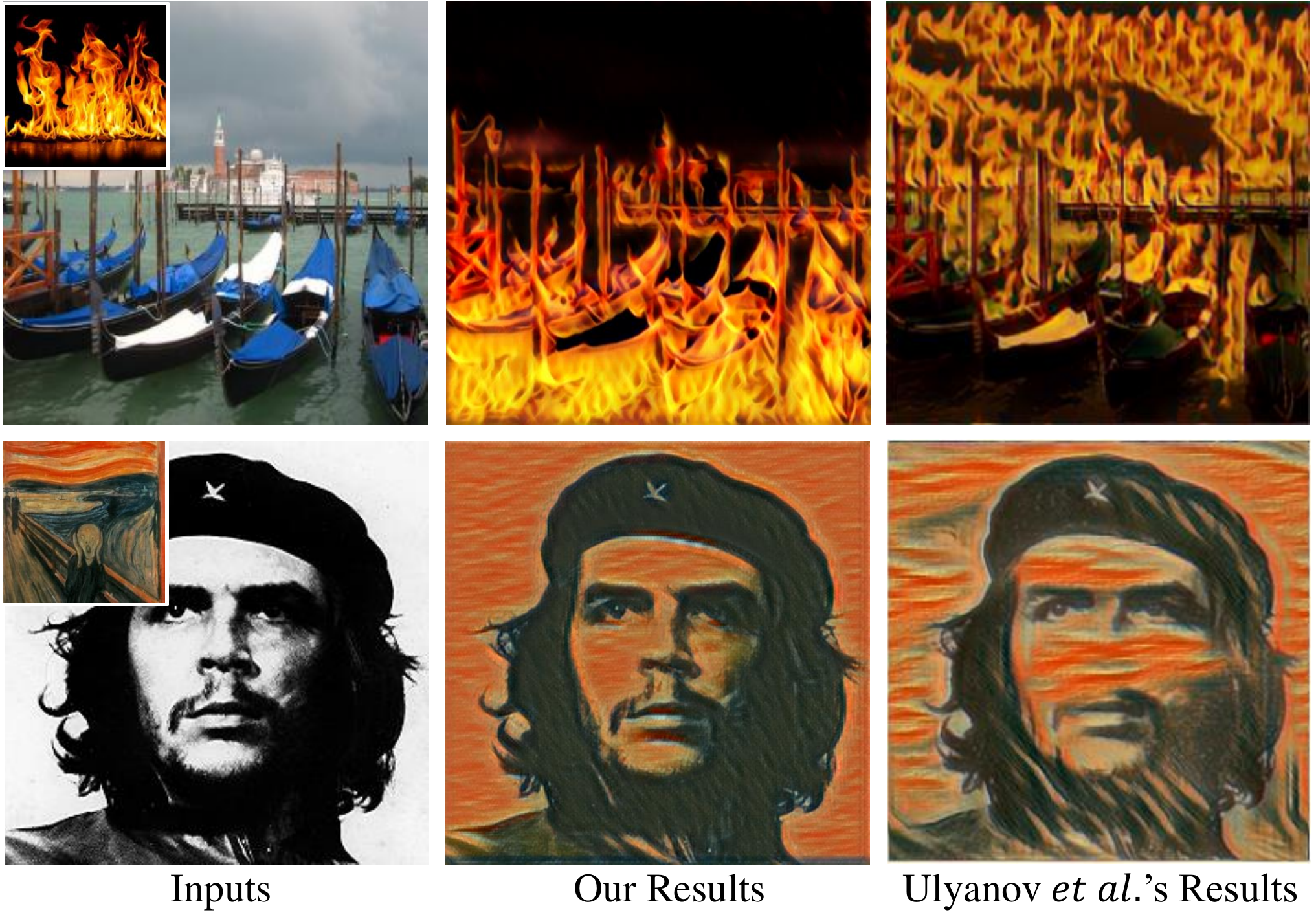}
		\caption{Comparison with the feed-forward network in \cite{ulyanov2016texture}.}
        \label{fg:comparison_texturnets}
\end{figure}
\begin{figure}
		\centering
		\includegraphics[width=\columnwidth]{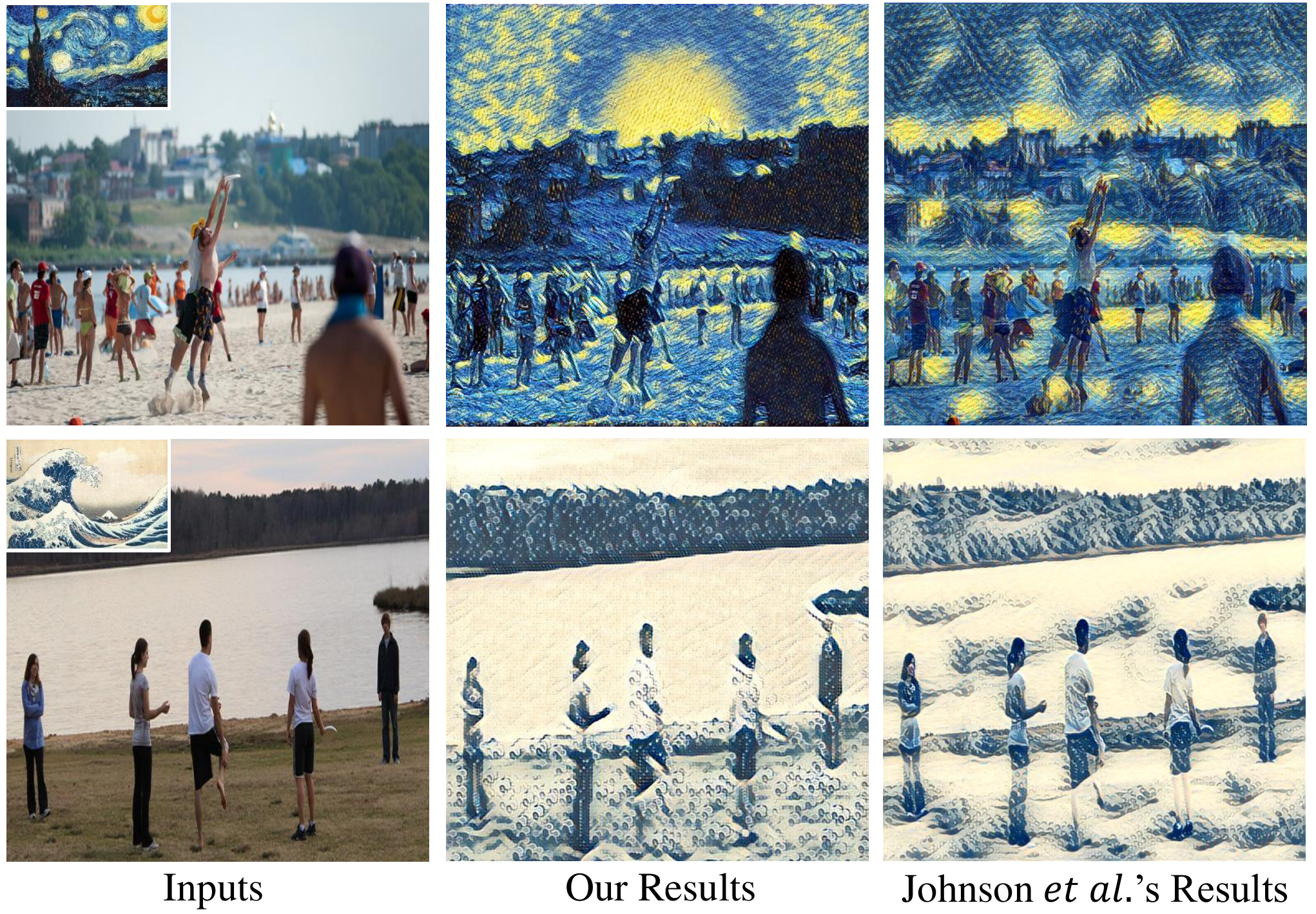}
		\caption{Comparison with the feed-forward network in \cite{johnson2016perceptual}.}
        \label{fg:comparison_johnson}
\end{figure}
In~\Fref{fg:comparison_texturnets} and~\Fref{fg:comparison_johnson}, we respectively compare our results with two feed-forward network methods \cite{ulyanov2016texture,johnson2016perceptual}.  We use $\alpha/\beta = 1/50$ (in~\Eref{eq:loss_perceptual}) in both comparisons. Ulyanov et al. \cite{ulyanov2016texture} design a shallow network specified for the texture synthesis task. When it is applied to style transfer task, the stylization results are more like texture transfer, sometimes randomly pasting textures to the content image. Johnson et al. \cite{johnson2016perceptual} use a much deeper network and often obtain better results. Compared with both methods, our results obviously present more region-based style transfer, for instance, the portrait in~\Fref{fg:comparison_texturnets}, and river/grass/forest in~\fref{fg:comparison_johnson}. Moreover, different from their one-network-per-style training, all of our styles are jointly trained in a single model.

\paragraph{Compared with other Synchronal Learning.}
\begin{figure}
		\centering
		\includegraphics[width=\columnwidth]{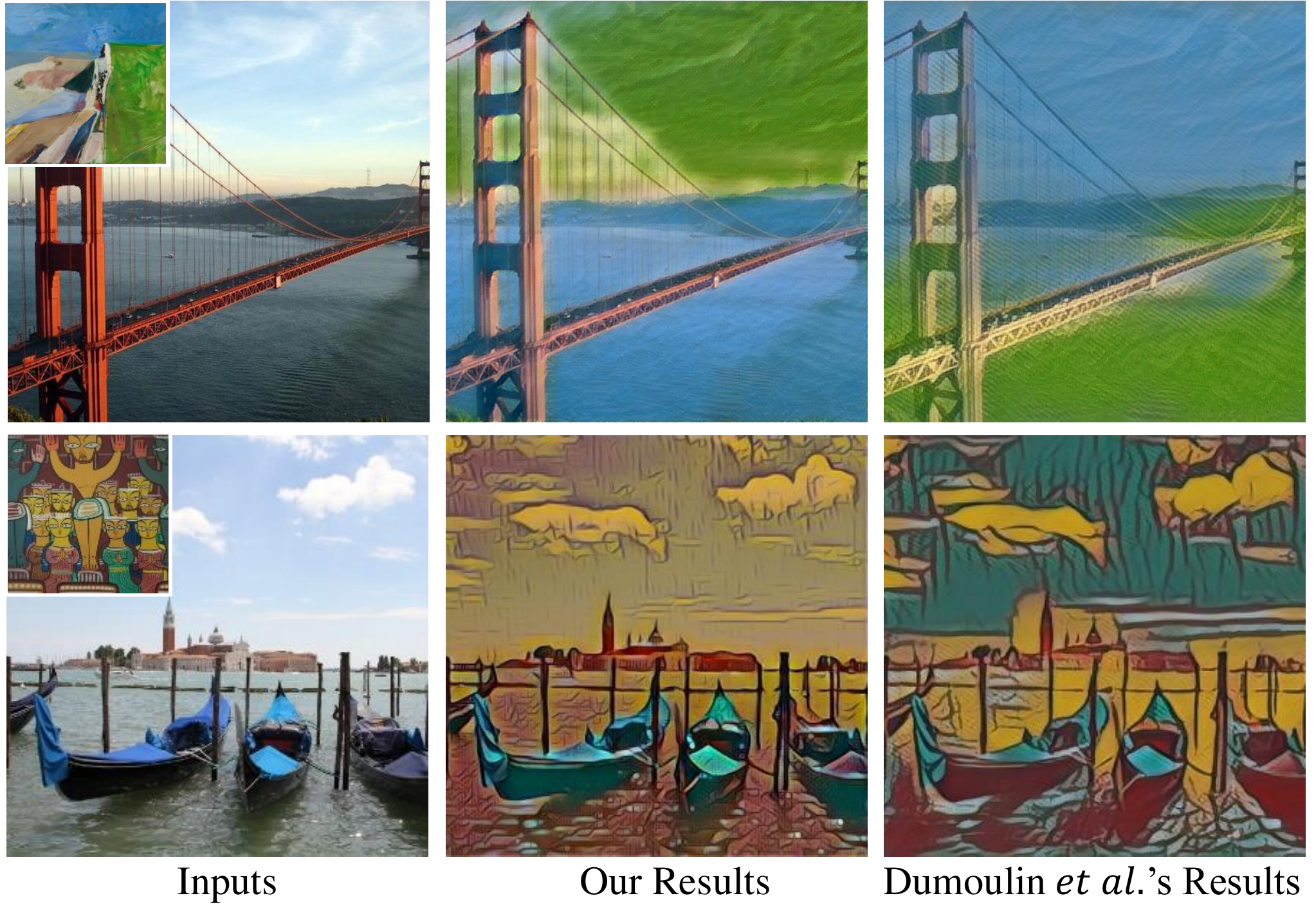}
		\caption{Comparison with the synchronal learning~\cite{2016arXiv161007629D}, }
        \label{fg:comparison_google}
\end{figure}
Dumoulin et al., in their very recent work \cite{2016arXiv161007629D}, introduces the ``conditional instance normalization" mechanism derived from \cite{ulyanov2016instance} to jointly train multiple styles in one model, where parameters of different styles are defined by different instance normalization factors (scaling and shifting) after each convolution layer. However, their network does not explicitly decouple the content and styles as ours. Compared with theirs, our method seems to allow more abilities of region-specific transfer. As shown in \fref{fg:comparison_google}, our stylization results better correspond to the natural regions of content images. In this comparison, we use $\alpha/\beta = 1/25$ (in~\Eref{eq:loss_perceptual}).

%------------------------------------------------------------------------
\section{Discussion and Conclusion}

In this paper, we have proposed a novel explicit representation for style and content, which can be well decoupled by our network. The decoupling allows faster training (for multiple styles, and new styles), and enables new interesting style fusion effects, like linear and region-specific style transfer. More importantly, we present a new interpretation to neutral style transfer which may inspire other understandings for image reconstruction, and restoration.

There are still some interesting issues for further investigation. For example, the auto-encoder may integrate semantic segmentation~\cite{long2015fully,Di2016ScribbleSup} as additional supervision in the region decomposition, which would help create more impressive region-specific transfer. Besides, our learnt representation does not fully utilize all channels, which may imply a more compact representation.

\section*{Acknowledgement}
This work is partially supported by National Natural Science Foundation of China(NSFC, NO.61371192)

{\small
\bibliographystyle{ieee}
\bibliography{egbib}
}

\end{document}